\documentclass{article}
\usepackage{amsmath, amssymb,relsize,bm,lmodern}
\usepackage{authblk}
\usepackage{graphicx}
\usepackage[T1]{fontenc}
\usepackage[utf8]{inputenc}

\newcommand{\R}{\mathbb{R}}

\begin{document}
\title{A scale-dependent notion of effective dimension}
\author[*]{Oksana Berezniuk}
\author[**]{Alessio Figalli}
\author[***]{Raffaele Ghigliazza}
\author[****]{Kharen Musaelian}
\affil[*]{\footnotesize CompatibL Sp, Prosta 32, 2nd floor, Warsaw, 00-838, Poland. E-mail: {\tt oksanab@compatibl.com}}
\affil[**]{\footnotesize ETH Z\"urich, Department of Mathematics, R\"amistrasse 101, 8092 Z\"urich, Switzerland. E-mail: {\tt alessio.figalli@math.ethz.ch}}
\affil[***]{\footnotesize ADQ Inc, 7 West 22nd St., 10th floor, New York, NY 10010, USA. E-mail: {\tt raffaele@adqinc.com}}
\affil[****]{\footnotesize Duality Group, 7 West 22nd St., 10th floor, New York, NY 10010, USA. E-mail: {\tt kharen@dualitygroup.com}}

\date{\today}
\maketitle

\begin{abstract}
We introduce a notion of ``effective dimension'' of a statistical model based on the number of cubes of size $1/\sqrt{n}$ needed to cover the model space when endowed with the Fisher Information Matrix as metric, $n$ being the number of observations. The number of observations fixes a natural scale or resolution. The effective dimension is then measured via the spectrum of the Fisher Information Matrix regularized using this natural scale.
\end{abstract}

A very important and challenging question in statistics and machine learning is the ``real'' dimension of a statistical model, such as a neural network.
Many definitions of effective dimension have been proposed in the literature, either based on the so-called VC dimension (see for instance \cite{Vapnik1998}), or on Gardner phase-space approach \cite{Opper1994}, or also on some effective dimension based on the rank of
the Jacobian matrix of the transformation between the parameters of the network and the parameters of the observable variables \cite{Geiger1996,Zhang2004} (see also \cite{Weigent1991,Bialek2001, Liang2017,Ravichandran2019}). Although these notions of dimension are all very natural when the number of observations go to infinity, they do not take into account the fact that only a finite-size sample of data is available.

The aim of this note is to propose a new definition of dimension that depends on the size of the data and that should give a better estimate on the dimension of the true model space that one observes in experiments. In other words, the size of the data fixes a natural scale/resolution at which one is able to observe the model, and such resolution influences the dimension.

\smallskip
Our notion is motivated by the theory of Minimum Description Length (MDL). We refer to the manuscript \cite{Grunwald2007} for an introduction to this important topic, and an exhaustive list of references.

Given a sample of $n$ data, and an effective
enumeration of models, MDL selects the model with the
shortest effective description that minimizes the sum of:\\
- the length, in bits, of an effective description of the model;\\
- the length, in bits, of an effective description of the data when
encoded with the help of the model.

Starting from this principle, given $(x_1,\ldots,x_n)=:x^n \in \mathcal X^n$ the space of possible $n$-data, 
and a statistical model $\mathcal M:=\{P(\cdot\,|\,\theta)\,:\,\theta \in \Theta\}$ for some $d$-dimensional parameter space $\Theta\subset \R^d$,
one defines the complexity (at size $n$)  of the model $\mathcal M$
as
$$
\mathbf{COMP}_n(\mathcal M):=\log\biggl(\sum_{x^n\in \mathcal X^n}P\bigl(x^n|\,\hat\theta(x^n)\bigr)\biggr),
$$
where $\hat\theta(x^n)\in \Theta$ is a maximizer of $\theta\mapsto P(x^n|\theta)$.

Let us assume that the model is i.i.d., so that one can define the Fisher Information Matrix $\bm{F}=(F_{ij})_{i,j=1}^d$ as
$$
{F}_{ij}(\theta):=\mathbb E\biggl[-\frac{\partial^2}{\partial \theta_i\partial\theta_j}\log P(X|\,\theta)\biggr],\qquad i,j \in \{1,\ldots,d\}.
$$
With this definition, and under suitable regularity conditions on $\mathcal M$ and $\Theta$, it is well known \cite{Rissanen1996,Takeuchi1997,Takeuchi1998,Takeuchi2000} that
$$
\mathbf{COMP}_n(\mathcal M)=\frac{d}{2}\log\frac{n}{2\pi}+\log\biggl(\int_{\Theta}\sqrt{\det \bm{F}(\theta)}\, d\theta\biggr)+o(1),
$$
where $o(1)\to 0$ as $n\to \infty.$

Usually, the term $d$ in the right hand side is interpreted as the dimension of the model, while the second term represents the geometric complexity of it.
Here, instead, we plan to combine these two terms to give a notion of \textit{effective dimension of the model $\mathcal M$ at scale $n$}.

\smallskip

Consider the Riemannian manifolds $(\Theta,\bm{g})$ endowed with the metric
$g_{ij}=\frac{n}{2\pi} F_{ij}$, where $F_{ij}$ is the Fisher Information.
Then 
$$
\frac{d}{2}\log\left(\frac{n}{2\pi}\right) +\log\left( \int_\Theta \sqrt{\det \bm{F}(\theta) }\, d\theta \right) 
=
\log\left(\int_\Theta \sqrt{\det \bm{g}(\theta)} \,d\theta \right).
$$
Note that $ \sqrt{\det \bm{g}(\theta)} \,d\theta $ is just the volume measure in Riemannian geometry.
Hence, in the formula above, we have taken the Riemannian manifold $(\Theta,\bm{F})$ endowed with the Fisher Information Matrix as metric, dilated the metric by a factor $\frac{n}{2\pi}$, and computed the logarithm of the volume of this manifold in this new metric.
Alternatively, if we think of the manifold $(\Theta,\bm{F})$ as a (isometrically embedded) subset of a larger Euclidean space, then $(\Theta,\bm{g})$ is the same as dilating $\Theta$ by $\sqrt{\frac{n}{2\pi}}$ while keeping the metric $\bm{F}$ constant. So, equivalently, we are considering the manifold $(\sqrt{\frac{n}{2\pi}}\Theta,\bm{F})$.

Now one would like to ask: what is the ``effective'' dimension of the manifold $(\Theta,\bm{g})$? 
Since we only have at our disposal $n$ observations, and  the optimal quantization of the parameter
space is achieved by using accuracy of order $\frac{1}{\sqrt{n}}$ (see \cite{Rissanen1978,Rissanen1996}), the idea is to use the box-counting dimension at scale $\frac{1}{\sqrt{n}}$.

\smallskip

Let us recall that the box-counting dimension (also called Minkowski dimension) is a way of determining the fractal dimension of a set $S$ in a Euclidean space  (see for instance \cite{Mandelbrot1982}).
This is defined by counting the number of boxes needed to cover the set on finer and finer grids. More precisely, 
if $N(\epsilon)$ is the number of boxes of side length $\epsilon$ required to cover the set, then the box-counting dimension is defined as
$$
{\rm dim}_{\rm box}(S) := \lim_{\epsilon \to 0} \frac {\log N(\epsilon)}{|\log \epsilon|}.
$$
Note that $N(\epsilon)$ is also equal to the number of boxes of side length $1$ required to cover the rescaled set  $\frac1{\epsilon}S$.

\smallskip

Motivated by this notion, we aim to define {\it effective dimension of $\mathcal M$ at scale $n$} as
$$
{\rm dim}_{{\rm eff},n}(\mathcal M)\approx \frac{\log\bigl(\#\{\text{cubes of size 1 needed to cover $\sqrt{\frac{n}{2\pi}}\Theta$\}}\bigr)}{\left|\log\sqrt{\frac{2\pi}{{n}}}\right|}
$$
(although not important for large $n$, for consistency with the previous formulas, we use $\sqrt{\frac{2\pi}{n}}$ in place of $\frac{1}{\sqrt{n}}$).

The  number of cubes of size 1 needed to cover $\sqrt{\frac{n}{2\pi}}\Theta$ can be thought as follows:
assume that $\Theta$ coincides with $[0,1]^d$, and that the matrix $\bm{F}$ is constant.
Also, for normalisation purposes, let us assume that the trace of $\bm{F}$ is equal to $d$:
$$
{{\rm tr}\bm{F}}=\sum_{i=1}^dF_{ii}=d.
$$
Diagonalize the matrix $\bm{F}$ as ${\rm diag}(s_1^2,\ldots,s_d^2)$.
Then, boxes of size $1$ with respect to the metric $\bm{F}$ are given by translations of the cube
$$
\biggl[0,\frac{1}{s_1}\biggr]\times \ldots\times \biggl[0,\frac{1}{s_d}\biggr],\qquad s_i>0,
$$
and the number of such cubes needed to cover $\sqrt{\frac{n}{2\pi}}[0,1]^d=\left[0,\sqrt{\frac{n}{2\pi}}\right]^d$ are equal to
$$
\prod_{i=1}^d\biggl \lceil\sqrt{\frac{n}{2\pi}}s_i\biggr\rceil\approx \sqrt{\prod_{i=1}^d\biggl(1+\frac{n}{2\pi}s_i^2\biggr)}=\sqrt{\det\left( {\rm Id}_d+\frac{n}{2\pi}\,\bm{F}\right)},
$$
where $\lceil s\rceil$ denotes the smallest integer greater than or equal to $s$, and ${\rm Id}_d$ is the identity matrix in $\R^{d\times d}$.

The intuition behind this argument is that, whenever ${{\rm tr}\bm{F}}$ is normalised to $d$, only the eigenvalues of $\bm{F}$ that are above ${\frac{2\pi}{n}}$ count in the definition of dimension. 
Motivated by this heuristic, we introduce the following:

\smallskip
\noindent
{\bf Definition:} {\it The effective dimension  of $\mathcal M$ at scale $n$ is given by the formula
\begin{equation}
\label{eq:def}
{\rm dim}_{{\rm eff},n}(\mathcal M):= 2\frac{\log\left(\mathlarger{\frac{1}{V_{\Theta}}\int}_\Theta \sqrt{\det\left( {\rm Id}_d+\frac{n}{2\pi}\,\hat{\bm{F}}(\theta)\right)}\,d\theta\right)}{\log\frac{n}{2\pi}},
\end{equation}
where $V_{\Theta} := \int_{\Theta}d\theta$ is the volume of $\Theta$, and $\hat {\bm{F}}=(\hat F_{ij})_{i,j=1}^d$ is defined as
$$
\hat F_{ij}:=d\,\frac{V_\Theta}{\int_{\Theta}{\rm tr}\bm{F}(\theta)\,d\theta}\,F_{ij}.
$$
}
\smallskip
Observe that the normalization of $\bm F$ by its averaged trace guarantees that $\frac{1}{V_{\Theta}}\int_\Theta{\rm tr}\hat {\bm{F}}(\theta)\,d\theta=d$, hence
 ${\rm dim}_{{\rm eff},n}(\mathcal M)$ is scaling invariant in $\theta$. Also, the term $\frac{1}{V_{\Theta}}$ ensures that, for $\hat {\bm F}$ constant, the notion of effective dimension is independent of the size of $\Theta$.

\smallskip
\noindent
{\it Remark 1.} It is important to notice that ${\rm dim}_{{\rm eff},n}$ may not be necessarily  increasing with respect to $n$. This is not surprising, since already in the standard definition of box-dimension there is no
monotonicity of the function  $\frac {\log N(\epsilon)}{|\log \epsilon|}$ with respect to $\epsilon.$ 
On the other hand, ${\rm dim}_{{\rm eff},n}(\mathcal M)$ is expected to capture more and more eigenvalues of $\hat {\bm F}$ as $n$ increases, and therefore to be monotonically increasing at macroscopic scales (in $n$).

\smallskip
\noindent
{\it Remark 2.}  Assuming non-degeneracy of the Fisher matrix, one can easily check that
\begin{equation}
\label{eq:limit}
{\rm dim}_{{\rm eff},n}(\mathcal M) \to d\qquad \text{as $n\to \infty$.}
\end{equation}
Note that the speed of this convergence is faster whenever the eigenvalues of $\hat {\bm F}$ are almost equal.
Indeed, in the extreme case $\hat {\bm F}={\rm Id}_d$, a Taylor expansion shows that
$$
{\rm dim}_{{\rm eff},n}(\mathcal M)=d+d\,\frac{\log\left(1+\frac{2\pi}{n}\right)}{\log \frac{n}{2\pi}}=d+O\biggl(\frac{d}{n\log n}\biggr).
$$
On the other hand, if the dispersion of eigenvalues of $\hat {\bm F}$ is high, then the small eigenvalues make the value of ${\rm dim}_{{\rm eff},n}(\mathcal M)$ decrease, and the convergence in \eqref{eq:limit} is expected to be slow.

\medskip

We conclude this note with some numerical simulations computing the relations between $n$, $d$, and ${\rm dim}_{{\rm eff},n}$, for some simple neural networks.
It turns out that, even for large $n$, the effective dimension is considerably smaller than $d$. As observed above, this is due to the high dispersion of eigenvalues, which makes the convergence in \eqref{eq:limit} slow in terms of $n$. Hence, ${\rm dim}_{{\rm eff},n}$ provides a much more effective bound with respect to $d$.

\begin{figure}[h]
\centering
\includegraphics[scale=0.52]{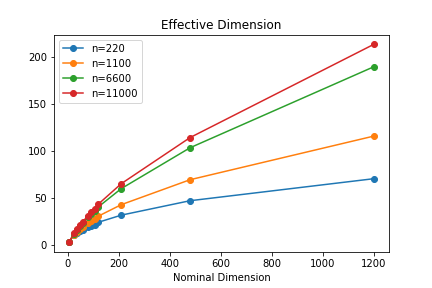}
\includegraphics[scale=0.52]{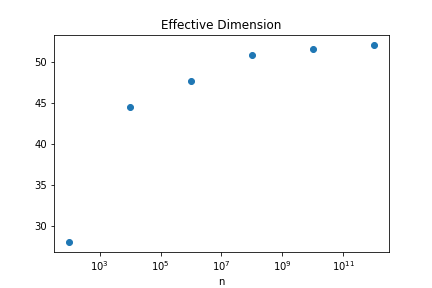}
\caption{In the first figure, for $n$ fixed, we increase $d$ (on the horizontal axis) and compute the effective dimension ${\rm dim}_{{\rm eff},n}(\mathcal M)$ (on the vertical axis).
Note that ${\rm dim}_{{\rm eff},n}$ is essentially increasing in $n$, and it is considerably smaller than $d$.
\newline
In the second figure, for $d=55$ we look how the effective dimension  (on the vertical axis) increases as a function of $n$
 (on the horizontal axis). As expected, ${\rm dim}_{{\rm eff},n}(\mathcal M) \sim d$ for $n$ large, but this requires very large $n$ with respect to $d$.}
\end{figure}

\bigskip

\noindent
{\it Acknowledgments.}
We thank Dario Villani, Ren\'e Carmona, Jonathan Kommemi, and Tomaso Poggio for useful discussions and comments.

\end{document}